\documentclass[11pt]{article}

\usepackage[preprint]{acl}

\usepackage{times}
\usepackage{latexsym}

\usepackage[T1]{fontenc}

\usepackage[utf8]{inputenc}

\usepackage{microtype}

\usepackage{inconsolata}

\usepackage{graphicx}

\usepackage[utf8]{inputenc} 
\usepackage[T1]{fontenc}    
\usepackage{hyperref}       
\usepackage{url}            
\usepackage{booktabs}       
\usepackage{amsfonts}       
\usepackage{nicefrac}       
\usepackage{microtype}      
\usepackage{xcolor}         
\usepackage{graphicx}
\usepackage{wrapfig}
\usepackage{amsmath}
\usepackage{natbib}
\usepackage{xspace}
\usepackage{amssymb}
\usepackage{float}
\usepackage[normalem]{ulem}
\usepackage{algorithm}
\usepackage{cleveref}
\usepackage{threeparttable}
\usepackage[normalem]{ulem}
\usepackage[most]{tcolorbox}
\usepackage{multirow}
\usepackage{makecell}
\usepackage{bm} 
\usepackage[most]{tcolorbox}
\tcbuselibrary{listingsutf8}
\usepackage{longtable}
\usepackage{array}
\usepackage{algpseudocode}

\newcommand{\eg}{e.g.,}

\newcommand{\fakeparagraph}[1]
{\vspace{0.5mm}\noindent\textbf{#1}}
\newcommand\seccref[1]{\S\ref{#1}}
\usepackage{caption}

\usepackage{amsmath,amsfonts,bm}

\def\eqref#1{equation~\ref{#1}}

\def\1{\bm{1}}

\def\vb{{\bm{b}}}

\def\vz{{\bm{z}}}

\def\mW{{\bm{W}}}

\DeclareMathAlphabet{\mathsfit}{\encodingdefault}{\sfdefault}{m}{sl}
\SetMathAlphabet{\mathsfit}{bold}{\encodingdefault}{\sfdefault}{bx}{n}

\def\sR{{\mathbb{R}}}

\newcommand{\modelname}{GrACE\xspace}
\newcommand{\concept}{GenConf\xspace}
\newcommand{\token}
{\texttt{<CNF>}\xspace}

\title{\modelname: A Generative Approach to Better Confidence Elicitation and Efficient Test-Time Scaling in Large Language Models}

\author{Zhaohan Zhang\textsuperscript{$1,\ast$}, Ziquan Liu\textsuperscript{$1$},  Ioannis Patras\textsuperscript{$1$}  \\
        \textsuperscript{1} Queen Mary University of London \\
        \textsuperscript{$\ast$} Corresponding author \\
        \texttt{
        \{zhaohan.zhang, ziquan.liu, i.patras\}@qmul.ac.uk
        } 
        }

\begin{document}
\maketitle
\begin{abstract}
Assessing the reliability of Large Language Models (LLMs) by confidence elicitation is a prominent approach to AI safety in high-stakes applications, such as healthcare and finance. 
Existing methods either require expensive computational overhead or suffer from poor calibration, making them impractical and unreliable for real-world deployment.
In this work, we propose \modelname, a Generative Approach to Confidence Elicitation that enables scalable and reliable confidence elicitation for LLMs. 
\modelname adopts a novel mechanism in which the model expresses confidence by the similarity between the last hidden state and the embedding of a special token appended to the vocabulary, in real-time.
We fine-tune the model for calibrating the confidence with targets associated with accuracy.
Extensive experiments show that the confidence produced by \modelname achieves the best discriminative capacity and calibration on open-ended generation tasks without resorting to additional sampling or an auxiliary model. 
Moreover, we propose two confidence-based strategies for test-time scaling with \modelname, which not only improve the accuracy of the final decision but also significantly reduce the number of required samples, highlighting its potential as a practical solution for deploying LLMs with reliable, on-the-fly confidence estimation.
  
\end{abstract}

\section{Introduction}

Large Language Models (LLMs) have become popular solutions to knowledge-intensive and logically complex problems for their capability of utilizing and reasoning over a huge amount of knowledge they learn \cite{achiam2023gpt, dubey2024llama, claude3,guo2025deepseek,yang2025qwen3}, especially in high-stakes domains such as healthcare \cite{singhal2023large}, law \cite{magesh2024hallucination}, and finance \cite{xie2024finben}.
Therefore, it is critical for LLMs to "know what they know" \cite{kadavath2022language} to become more reliable and gain users' trust in their deployment.

\begin{figure}[t]
    \centering
    \resizebox{0.48\textwidth}{!}{    \includegraphics{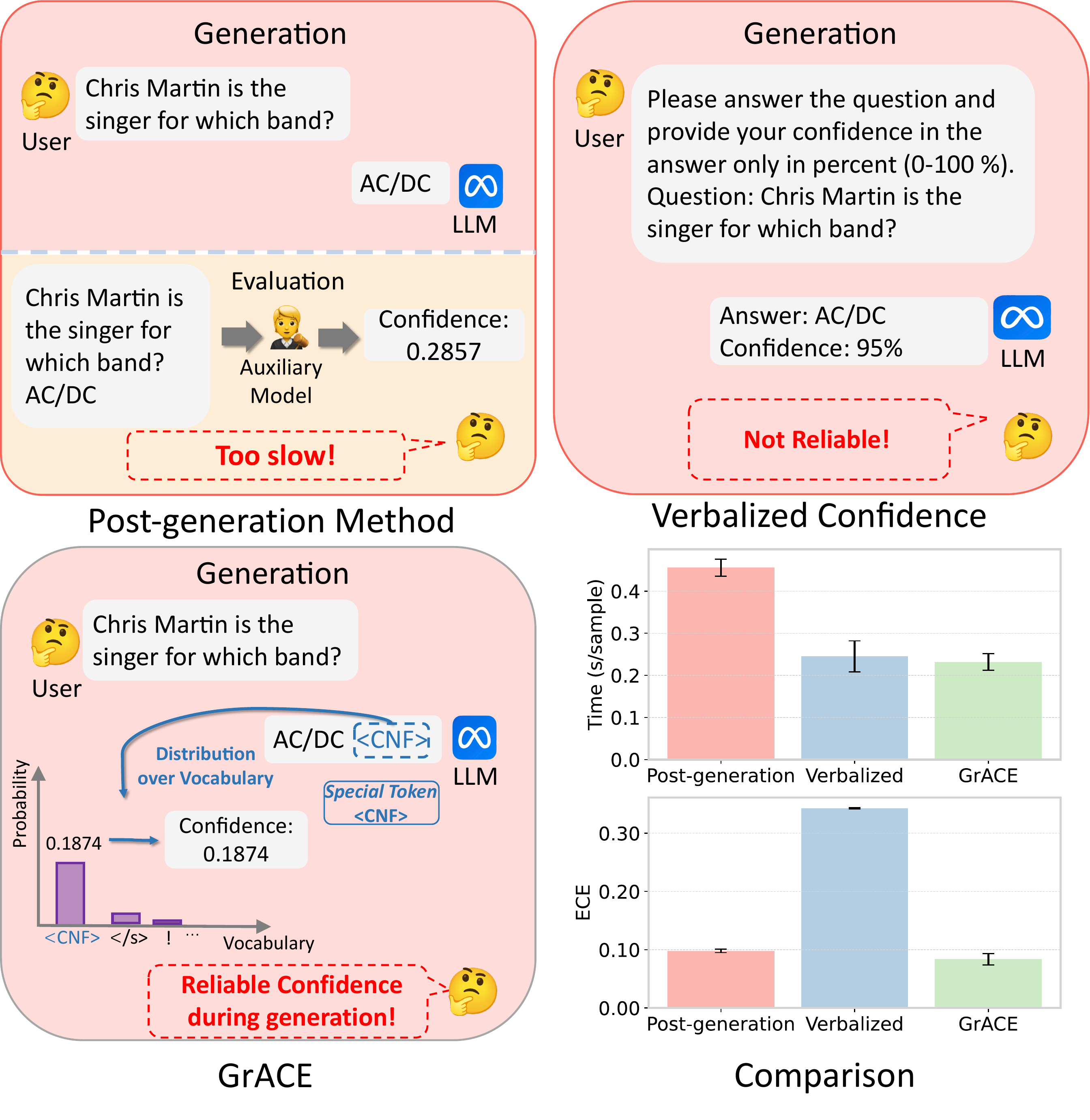} 
    }
	\caption{
    \textbf{Comparison among the workflow and performance of post-generation methods, verbalized confidence, and \modelname.}
\protect\footnotemark.
	}
    
    \label{fig:intro}
\vspace{-0.5cm}
\end{figure}
\footnotetext{We report time and Expected Calibration Error (ECE) using two representative methods in post-generation methods and verbalized confidence, namely \textit{P}(True) \cite{kadavath2022language} and verbalized confidence \cite{tian2023just}.
The experiment is conducted with the Llama2-7B model on the TriviaQA dataset, using a single NVIDIA GeForce RTX 3090 GPU.}

Estimating model confidence is a promising approach to assessing the reliability of model predictions.
A desirable confidence score should exhibit strong separability between positive and negative samples (discrimination) and align closely with actual correctness (calibration).
Although confidence elicitation and calibration have been extensively studied in classification tasks \cite{guo2017calibration, desai2020calibration, li2025large}, recent attention has shifted to open-ended generation tasks (e.g., question answering) for LLMs.
Nevertheless, as illustrated in Figure \ref{fig:intro}, existing methods exhibit limitations in \textit{when} a model expresses confidence and \textit{how well} that confidence is calibrated, which constrains their practical utility in applications such as Test-Time Scaling (TTS) \cite{brown2024large}.
Post-generation methods, which assign the precise confidence score after generation, dominate current confidence elicitation approaches.
They elicit confidence either by prompting the original model again \cite{kadavath2022language, zhang2024atomic} or using auxiliary models \cite{ulmer2024calibrating, kapoorlarge}.
Such methods require an additional confidence evaluation stage after generation, increasing latency and degrading human-computer interaction.
Verbalized confidence offers a generative alternative by producing numerical values \cite{tian2023just, xiongcan} or linguistic expressions (\eg{most likely, likely, not sure}) \cite{band2024linguistic, yang2024logu, yang2025uncle} during the generation stage. 
However, verbal expressions of confidence are inherently vague as the discrete vocabulary fails to capture nuance differences in confidence  -- for example, the confidence intervals for "most likely", "likely", and "unlikely" are neither standardized nor quantifiable.
Moreover, LLMs tend to be overconfident when expressing verbal confidence, potentially due to too much focus on imitating human expressions \cite{xiongcan}.

In this paper, we enable LLMs to learn the true distribution of the prediction correctness and express the confidence with \textbf{G}ene\textbf{r}ative c\textbf{A}librated \textbf{C}onfidence \textbf{E}licitation (GrACE), making it an efficient and reliable alternative for existing methods.
\modelname adds a special token \token to the vocabulary and teaches the model to generate it at the end of the response for confidence elicitation.
Specifically, we fine-tune the model to express confidence using the similarity between the last hidden state and the embedding of \token, leveraging the inherent monotonicity and simplicity of the similarity function.
To ensure the elicited confidence is discriminative and calibrated, we construct a calibration set where training samples are grouped according to the internal confidence of the model \cite{liu2024enhancing,ji2025calibrating} and assign group accuracy as the calibration target.
To reduce the cost of fine-tuning, we only fine-tunes LoRA \cite{hulora} and the embedding of \token 
when implementing \modelname.
The fine-tuning cost is amortised by the learned scalable confidence elicitation mechanism that generalizes across prompts and domains without an additional evaluation stage at test time.
The on-the-fly confidence elicitation mechanism, together with discriminative and reliable confidence scores, makes \modelname naturally well-suited for applications such as TTS, which demand dependable decision-making and real-time responses.

Our contributions are summarized as follows:
\begin{itemize}
    \item We propose a generative confidence elicitation method \modelname, which provides a novel and scalable mechanism for discriminative and calibrated confidence elicitation on-the-fly with negligible computation overhead. 
    \item We design two strategies that integrate the real-time confidence scores generated by \modelname into the aggregation and adaptive sampling process of test-time scaling, improving both efficacy and efficiency.
    
\item Extensive experiments on open-ended generation tasks and test-time scaling demonstrate that \modelname not only produces more accurate confidence estimates with strong generalization to unseen domains, surpassing the state-of-the-art method (Apricot \cite{ulmer2024calibrating}) by 2.6\%, 1.8\%, and 4.4\% in terms of ECE, Brier score, and AUROC, respectively, but also provides reliable guidance that improves accuracy by an average of 1.54\% while substantially reducing the number of samples required during test-time scaling.


\end{itemize}

\section{Related Works}
\fakeparagraph{Post-generation Confidence Elicitation.}
Unlike model uncertainty, which captures the variability in a model’s responses to questions \cite{kuhnsemantic, zhang2025get}, confidence elicitation directly assigns an explicit confidence score to the model’s prediction.
Post-generation methods follow the generate-then-evaluate paradigm, which first acquires model generations, and then evaluates confidence score with an auxilarity model.
\citet{kadavath2022language} proposes \textit{P}(True), which asks the model to self-evaluate its confidence on generation by converting the open-ended task to the multiple choice setting with careful prompting.
Further works \cite{kapoorlarge} calibrate \textit{P}(True) by fine-tuning the model with LoRA using binary labeled data.
As for more efficient solutions 
\cite{azaria2023internal, kossen2024semantic, liu2024enhancing}, confidence can be detected from the model's hidden state with a linear probe.
Post-generation methods are also applicable to black / gray-box models.
\citet{ulmer2024calibrating} trains an auxiliary model with semantic-aware targets which predicts model confidence based on generation only.
\modelname differs from post-generation methods by eliminating the separate evaluation stage and producing the confidence score in the generation stage, thereby reducing computational overhead and accelerating the confidence elicitation process.

\fakeparagraph{Verbalized Confidence Elicitation.}
Another branch of confidence elicitation aims to teach the models to express their confidence in natural language.
\citet{tian2023just} designs different prompting strategies to lead the model to output a calibrated confidence score.
Some works \cite{band2024linguistic, yang2024logu, hager2025uncertainty} endow the model with the ability to use confidence-related phrases (e.g., not sure, uncertain) to express confidence as humans do.
\citet{cohen2024don} asks the model to generate a special token \texttt{[idk]} when it feels uncertain about the responses.
Although verbalized confidence is a natural and efficient approach to eliciting model confidence, LLMs are found to be overconfident when verbalizing their confidence \cite{xiongcan}.
Moreover, the confidence in the form of phrases conveys vague and unquantifiable information, restricting its application to broad topics.

\fakeparagraph{Test-Time Scaling.}
Test-Time Scaling (TTS) \cite{snell2024scaling} improves the response quality of LLMs with additional computing during the inference stage.
Repeated sampling \cite{brown2024large} with decision-making rules such as Self-Consistency \cite{wang2022self} is a common practice for selecting high-quality samples for given prompts.
However, this process requires a fixed sample size during TTS, which is not always necessary \cite{chen2024not}.
While some works investigate optimal computational resources allocation \cite{snell2024scaling, muennighoff2025s1}, there is research focusing on dynamically adjusting the sample size.
Some attempts \cite{liescape,aggarwal2023let} determine when to terminate sampling by pre-sampling and calculating statistical features within the pre-sample set.
As a metric to evaluate the response quality, confidence is gradually applied to the decision-making process for improving TTS.
\citet{taubenfeld2025confidence, huang2025efficient} optimize the sample size through the lens of confidence and calibration.
However, these works are based on \textit{P}(True) which requires an additional evaluation stage as previously noted.
Concurrent work \cite{fu2025deep} leverages confidence to filter out low-quality reasoning traces but does not account for calibration, resulting in threshold selection without risk guarantees.
Our work enables on-the-fly elicitation of discriminative and well-calibrated confidence, supporting efficient, effective, and risk-controlled TTS.

\section{Methodology}
In this section, we introduce the formulation of the generative confidence elicitation problem in \seccref{sec:probform} and the \modelname framework in \seccref{sec: method}, including the form of confidence elicitation in \modelname \seccref{subsec:confidence}, the calibration target \seccref{subsec:cali_target}, training process \seccref{subsec:train}, and \modelname's application on test-time scaling \seccref{subsec:tts}.

\subsection{Problem Formulation}
\label{sec:probform}
Given a prompt $\bm{x}$ and a model $\mathcal{M}$, we define the generative confidence (\concept) $c$ as:
\begin{equation}
    (\bm{y}_i, \vz, c_{i}) = \mathcal{M}(\bm{x}), \ 
    c_{\text{i}} \in \sR^+,
\label{eq1}
\end{equation}
where $\bm{y}_i=\{y_i^1, ..., y_i^N\}$ is the $i$-th sample, $\vz=\{z_1, ..., z_N\}$ is the model's hidden state, $c_i$ is the confidence score corresponding to the sample $\bm{y}_i$.
Note that Eq.\ref{eq1} requires no auxiliary model and the confidence $c$ should be a real number extracted during generation rather than a textual string.
To be effective and trustworthy in downstream applications, the \concept $c$ should exhibit both discrimination and calibration.
Formally:
\begin{align}
& \quad \mathbb{P}(c_m>c_n|f(\bm{y}_m)=1, f(\bm{y}_n)=0)=1, \\
& \quad \mathbb{P}(f(\bm{y})=1 | \bm{x}, \bm{y}, c) = c,   
\end{align}
where $f(\cdot)$ is a function which assigns 1 to the correct response and 0 to the wrong response.
Given these notions, discrimination requires 
$c$ to correctly rank positive samples above negative ones, whereas calibration requires $c$ to reflect the actual accuracy of the model’s responses.

\subsection{\modelname}
\label{sec: method}
\begin{figure*}[htbp]
    \centering
    \resizebox{0.8\textwidth}{!}{%
        \includegraphics{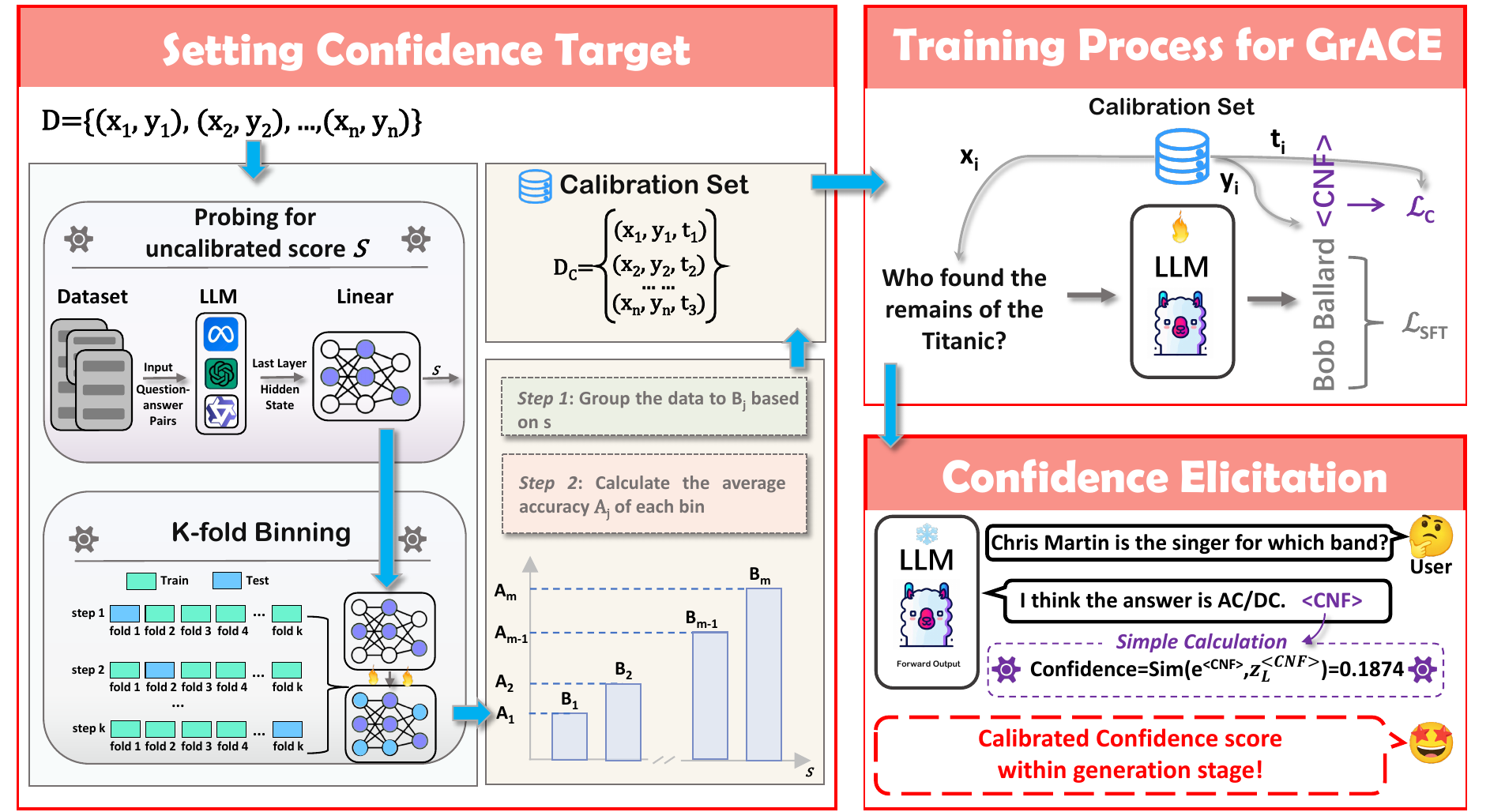} 
    }

    \caption{\textbf{The illustration of \modelname framework.} 
    We group the calibration data using the model's self-awareness and use the accuracy of each group as the calibration target (\seccref{subsec:cali_target}).
    The model is trained with the combination of calibration loss and supervised fine-tuning loss (\seccref{subsec:train}).
    The confidence is elicited by the distance between the hidden state and \token embedding (\seccref{subsec:confidence}).}
    \label{fig:method} 
    \vspace{-0.5cm}
\end{figure*}

We describe the \modelname framework (illustrated in Figure \ref{fig:method}) for teaching the model to produce \concept in this section.
\subsubsection{Confidence Elicitation in \modelname}
\label{subsec:confidence}
Ideally, the confidence should be expressed in continuous space during generation as proposed in \seccref{sec:probform}.
Inspired by the pause token mechanism \cite{goyalthink, wangguiding} which prompts the model to "think" by generating a sequence of special, non-semantic tokens before reasoning, we introduce a new token, \token, into the vocabulary to encourage the reflective behavior of the model in the latent space.
Specifically, we append \token to the end of each question-answer pair.
We train the model to generate \token at the end of its response, thereby enabling an additional forward pass in which the model can reflect on its output and estimate confidence in the latent space.
In what follows, we describe how the model expresses its confidence in the latent space.

Let $\bm{e}^{\token}$ be the embedding of the new token \token, and $\bm{z}_L^{\token}$ be hidden states in the last layer $L$ corresponding to the \token.
Then, the confidence is estimated as the similarity between $\bm{e}^{\token}$ and $\bm{z}_L^{\token}$, that is:
\begin{equation}
    c = \mathrm{sim}(\bm{z}_L^{\token}, \bm{e}^{\token}).
\end{equation}
In causal language models, the language modeling head (typically a linear projection followed by a softmax) serves as a mechanism to project the last hidden state to the vocabulary.
This projection implicitly evaluates the similarity between the last hidden state and each token embedding in the vocabulary.
We leverage this built-in capability as the foundation for defining function $\mathrm{sim}(\cdot)$, aligning it with the model's native representation geometry:
\begin{equation}
    c = \mathrm{softmax}(\bm{E}\bm{z}_L^{\token})_{\token},
\end{equation}
where $\bm{E} \in \mathbb{R}^{(|\mathcal{V}|+1)	\times d}$ is the embedding matrix of the vocabulary, $d$ is the dimension of token embeddings, the subscript \token is the position associated with \token.
Note that the position of \token is flexible depending on the segment that the users care about, we provide the experiment results with \token prepending the response in Ablation Study in Appendix \ref{appd:ablation}.
\subsubsection{Setting Target Confidence for \modelname} 
\label{subsec:cali_target}
To guide the model to generate a discriminative and well-calibrated confidence score $c$, it is important to set target confidences for responses $\bm{y}$ that are distributionally sharp and numerically aligned with their actual probability of correctness \cite{kuleshov2015calibrated}.
Traditional methods use binary labels as confidence targets, leading to only a sharp distribution with poor calibration \cite{niculescu2005predicting}.
We propose to smooth the confidence target and ensure calibration by grouping the training data and assigning the empirical accuracy within each group as the target confidence.
The grouping is determined by the outputs of a linear probe trained with binary labels, which helps ensure a sharper distribution. 
Specifically, given training set $
\mathcal{D} = \left\{ (\bm{x}_i, \bm{y}_i, f(\bm{y}_i)) \right\}_{i=1}^{|\mathcal{D}|}$,
we probe an uncalibrated confidence score $s$ from the last hidden layer $L$ of the model for generation in sequence length $N$ with a linear probe parameterized with $\theta$, formulated as:
\begin{equation}
\begin{aligned}
\small
    s = \mW_{\theta}^T&\bar{\bm{z}}_L + \vb_{\theta}, \quad
    \bar{\bm{z}}_L = \frac{1}{N}\sum_{i=1}^{N}\bm{z}^i_{L}
\end{aligned}
\end{equation}
We perform k-fold binning \cite{liu2024enhancing} to align $s$ from probing with response accuracy for every training sample.
Specifically, we partition the dataset into $k$ folds, iteratively train the linear probe on $k-1$ folds with mean squared error loss, and test on the remaining fold until all the samples in the training data $(\bm{x}, \bm{y})$ are annotated with $s$.
To convert $s$ to response accuracy, we group the data into $M$ bins with equal intervals according to $s$ and calculate the accuracy $\text{A}_m$ in each bin $m$.
Let the target $t_i = A_m$ if the data point $i\in\mathcal{B}_m$, where $\mathcal{B}_m$ denotes the set of indices belonging to bin $m$. 
The resulting calibration set is defined as $
\mathcal{D}_c = \left\{ (\bm{x}_i, \bm{y}_i, t_i) \right\}_{i=1}^{|\mathcal{D}_c|}$.
In this way, we obtain targets for \modelname that are numerically equal to the accuracies within each group and sharply distributed, as the probe is trained to assess the factuality of the question–answer pairs.
In Appendix~\ref{appd:target} and~\ref{appd:relationship}, we plot the kernel density estimation of the target confidence and demonstrate that aligning the predicted confidence with the target confidence 
$t$ improves the calibration by reducing the expected calibration error.

\subsubsection{Training Process}
\label{subsec:train}
We optimize the model to align its self-reflective confidence $c$ with the calibration target $t$ using the Mean Squared Error (MSE) as calibration loss.
However, fine-tuning the model towards only the calibration objective will sacrifice the model's overall performance.
To maintain the model utility, we adopt the supervised fine-tuning (SFT) loss on the original generation of the model as a regularizer.
Note that we exclude the \token from the calculation of SFT loss to avoid the conflict between the two objectives. 
The total loss is formulated as:
\begin{equation}
\resizebox{\columnwidth}{!}{$
\begin{aligned}
\mathcal{L}_{T}
   &= \mathcal{L}_{\text{C}}
     + \gamma \mathcal{L}_{\text{SFT}} \\
   &= \frac{1}{|\mathcal{D}_c|}
      \sum_{i=1}^{|\mathcal{D}_c|}
        \bigl(t_i - \mathrm{sim}(\bm{z}_{L}^{\token}, e^{\token})\bigr)^2
      - \gamma \log p_{\theta}(\bm{y}_i \mid \bm{x}_i)
\end{aligned}
$}
\label{eq:loss}
\end{equation}
where $\gamma$ is a hyperparameter for balancing the loss terms $\mathcal{L}_{\text{C}}$ and $\mathcal{L}_{\text{SFT}}$. 
This objective prioritizes the token \token at the target position by assigning it a higher probability while suppressing competing tokens, thereby training the model to consistently produce \token at the end of its response.

\subsubsection{Test-Time Scaling with \modelname}
\label{subsec:tts}

As mentioned in \seccref{sec:probform}, \modelname helps the model to output discriminative and well-calibrated confidence with negligible inference cost.
Building upon this, we incorporate \modelname into the common aggregation and adaptive sampling practices, namely self-consistency \cite{wang2022self} and early-stopping \cite{liescape}, in TTS for improving the scaling law at inference time without much additional computation for confidence estimation.

\fakeparagraph{Self-Consistency with \modelname (\modelname-SC).}
Self-consistency determines the accepted response by taking a majority vote over all samples generated through TTS.
Intuitively, given the variance in sample quality, the voting process should account for this heterogeneity by assigning greater weight to samples with higher confidence.
Thus, we leverage the discrimination of the confidence and reformulate the self-consistency.
Given the sample answer set $\mathcal{A}=\{ a_1, ..., a_T\}$ where the answer $a_i$ is the final answer extracted from model response $\bm{y}_i$ and $T$ is the sampling budget, \modelname-SC determines the final answer $a^*_{sc}$:
\begin{equation}
    a^*_{sc} = \arg\max_{a \in \mathcal{A}} \sum_{i=1}^{T} \mathbb{I}(a_i = a) \cdot c_i
    ,
\label{eq:sc}
\end{equation}
where $\mathbb{I}(\cdot)$ is an indicator function which returns 1 if the answer extracted from the $i$-th sample $a_i$ matches $a$ else returns 0.

\begin{table*}[!t]
\centering
\renewcommand\arraystretch{0.9}
\resizebox{\textwidth}{!}{
\begin{tabular}{cc|cccc|cccc}
\toprule \midrule
        &  \multicolumn{1}{c}{}    & \multicolumn{4}{c}{\textbf{TriviaQA}}                                                           & \multicolumn{4}{c}{\textbf{SciQ}}                                                              \\ \cmidrule(lr){3-6} \cmidrule(lr){7-10}
\textbf{\textit{Backbone}}& \multicolumn{1}{c}{\textbf{\textit{Methods}}}       & ECE $\downarrow$ & Brier $\downarrow$ & \multicolumn{1}{c}{ AUROC $\uparrow$}& Acc$\uparrow$ & ECE $\downarrow$ & Brier $\downarrow$ & AUROC $\uparrow$ & Acc$\uparrow$ \\ \midrule
\multirow{7}{*}{\makecell[c]{Phi3-3.8B }} &
Seq. Likelihood & 14.57                     & 20.62                       & \underline{81.45}  & 64.40                    & 14.47                   & 23.81               & 73.69   & 58.50            \\ 
&Platt Scaling       & 10.75                 & 22.43                      & \underline{81.45} & -                       & 10.83                      & 24.57                     & 73.69   & -               \\ 
&P(True)   & 12.65                       & \underline{17.85}                 & 81.69   & -            & 31.82                & 30.09                     & \underline{79.66}   & -             \\ 
&Verbal    & 27.34             & 28.96             & 63.01  & -            & 36.52             & 36.55             & 56.44  & -           \\ 
&Apricot    & 9.69              & \underline{17.85}              & 78.36     & -       & \underline{7.09}             & \underline{18.80}             & 76.45   & -         \\
&ActCab    & \underline{8.47}              & 22.46              & 64.27    & -          & 8.72             & 22.20             & 70.24  & -           \\
&\modelname     & \textbf{7.63}              & \textbf{16.64}              & \textbf{81.85}    & 64.70          & \textbf{5.21}             & \textbf{16.93}   &    \textbf{82.37}       & 58.10      \\
\midrule

\multirow{7}{*}{\makecell[c]{Llama2-7B}} &
Seq. Likelihood
& 12.56                      & 21.88                       & 73.27 & 59.90                    & 16.52                   & 26.74                & 61.41  & 57.20                   \\ 
&Platt Scaling
&  13.95               & 23.47                      & 73.27     & -                 & 10.12                      & 24.84                      & 61.41     & -                \\ 
&P(True)
& 9.82                     & 23.12                & 66.02    & -             & \underline{7.46}               & 24.94                     & 49.41     & -         \\ 
&Verbal     & 34.31              & 37.02             & 50.89    & -         & 56.05            & 37.94           & 53.07    & -       \\ 
&Apricot     & 11.88            & \underline{18.55}             & \underline{81.61}     & -       & 8.42            & \underline{16.69}           & \underline{82.04}     & -       \\
&ActCab     & \textbf{5.18}             & 23.75             & 56.21    & -         & 9.34             & 24.51             & 60.00 & -            \\
&\modelname     & \underline{8.41}             & \textbf{16.78}             & \textbf{84.22}    & 59.10        & \textbf{5.45}          & \textbf{14.33}            & \textbf{87.72}     & 57.40     \\
\midrule

\multirow{7}{*}{\makecell[c]{Llama3.1-8B}} &
Seq. Likelihood
& 15.71                      & 17.32                       & \underline{79.44} & 78.00                    & 24.63                   & 27.26                & 69.46  & 62.80                   \\ 
&Platt Scaling
&  13.26               & 16.96                      & \underline{79.44}     & -                 & \underline{9.03}                      & 23.37                      & 69.46     & -                \\ 
&P(True)
& 17.33                     & 22.24                & 53.73    & -             & 17.96               & 27.90                     & 52.83     & -         \\ 
&Verbal     & 11.45              & 17.36             & 69.96    & -         & 21.11            & 26.20           & 64.09    & -       \\ 
&Apricot     & 9.80            & \underline{15.56}             & 74.44     & -       & 17.06            & 24.17           & \underline{69.94}     & -       \\
&ActCab     & \underline{7.84}             & 16.01             & 72.74    & -         & 10.99             & \underline{21.65}             & 69.12 & -            \\
&\modelname     & \textbf{5.94}             & \textbf{12.95}             & \textbf{81.03}    & 78.40        & \textbf{8.16}          & \textbf{19.13}            & \textbf{75.62}     & 63.50     \\
\midrule
\bottomrule
\end{tabular}
}
\vspace{-0.2cm}
\caption{\textbf{Performance comparison for open-end generation calibration task.} 
Note that the competitors do not fine-tune the backbone model so the accuracy is the same among the competitors.
We report all the results in percentage (\%).
The best results are \textbf{bolded}.
}
\label{Tab: Calibration}
\vspace{-0.5cm}
\end{table*}

\fakeparagraph{Early Stopping with \modelname (\modelname-ES).}
Early stopping terminates the sampling process once the generated samples meet predefined criteria, thereby reducing the number of samples required by the TTS algorithm.
Recent approaches \cite{liescape, wan2025reasoningawareselfconsistencyleveraging, aggarwal2023let} build upon the concept of self-consistency.
They partition the sampling trajectory into smaller windows and halt further sampling once the consistency within a window exceeds a specified threshold.
However, these methods typically rely on pre-sampling to compute the stopping criterion.
As a sampling-free method, \modelname produces discriminative and calibrated confidence scores in real-time for each sample during inference, which serves as an effective and reliable criterion for early-stopping.
We keep sampling until we obtain a sample with confidence larger than the threshold $\tau$ and use this sample as the final prediction.
If the sampling budget runs out and no sample's confidence reach the threshold, we pick the final answer using \modelname-SC.
The pseudocode for \modelname-ES is in Appendix \ref{appd:code}.

\vspace{-0.1cm}
\section{Experiments}
\label{sec:exp}
We introduce the experimental settings and the results in this section to demonstrate the effectiveness of \modelname on both 1) open-ended generation and 2) test-time scaling.
\vspace{-0.1cm}
\subsection{\modelname for Open-ended Generation}
\vspace{-0.2cm}
\label{sec: openend}
\begin{figure}[t]
    \centering
    \vspace{-0.2cm}
    \resizebox{0.48\textwidth}{!}{    \includegraphics{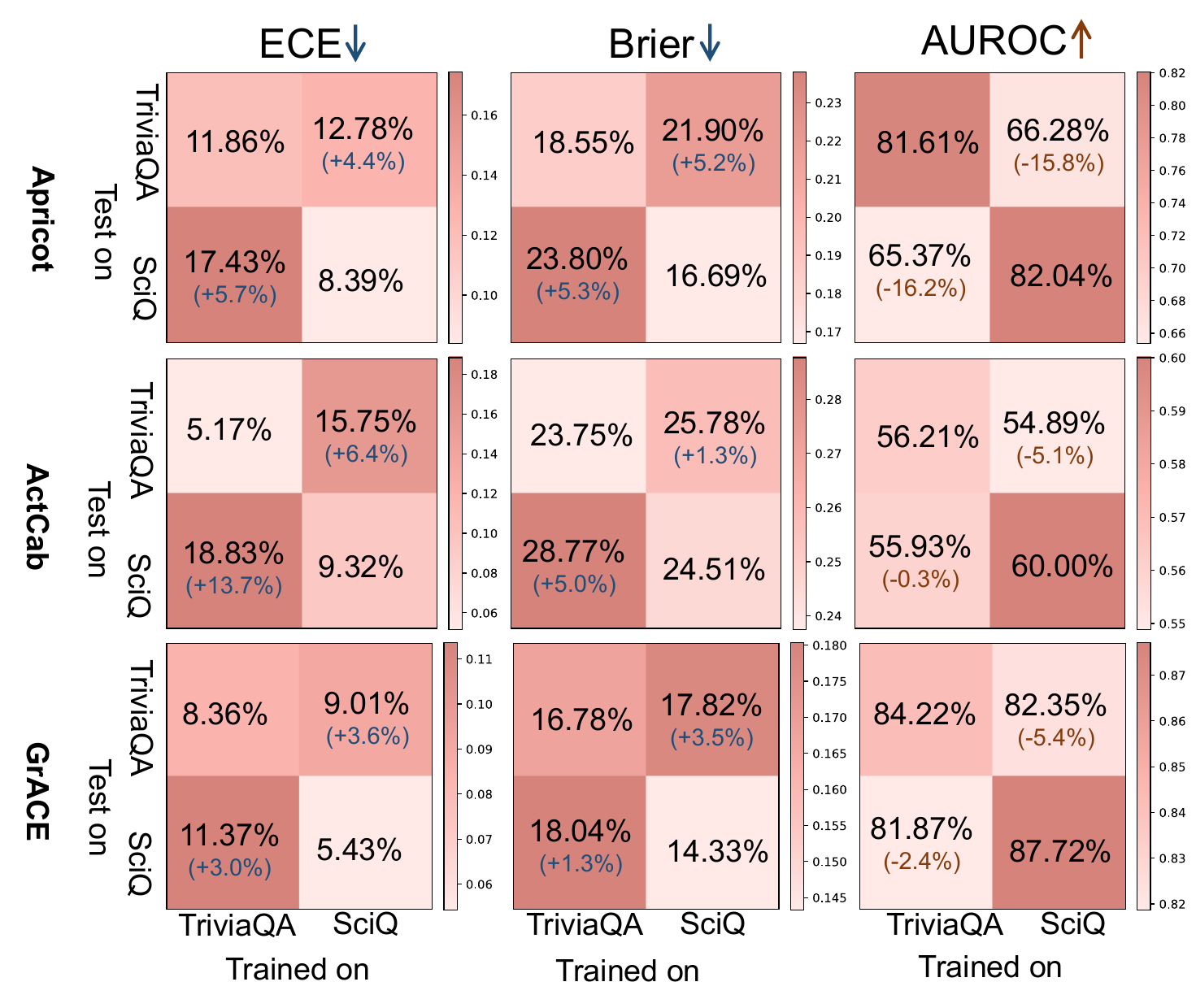} 
    }
	\caption{
    \textbf{Generalization ability of different confidence elicitation methods.} 
        The values in the brackets represent how much the model’s performance changes when moving from its training domain to an unseen domain.
	}
    
    \label{fig:generalization}
    \vspace{-0.5cm}
\end{figure}

\begin{figure*}[t]
  \centering
  \vspace{-1em}
  \noindent
  \begin{minipage}[t]{0.5\textwidth}
    \centering
    \captionsetup{type=table}
    \resizebox{\linewidth}{!}{
      \begin{tabular}{cc|cc|cc}
        \toprule \midrule
        \multirow{3}{*}{\textbf{Backbone}}& \multirow{3}{*}{\textbf{Method}}& \multicolumn{2}{c}{\textbf{MathQA}} & \multicolumn{2}{c}{\textbf{ARC\_C}} \\
        \cmidrule(lr){3-4} \cmidrule(lr){5-6}
         &  & Acc $\uparrow$ & $\hat{T}$ $\downarrow$ & Acc $\uparrow$ & $\hat{T}$ $\downarrow$ \\
        \midrule
        \multirow{5}{*}{Llama-3.1} &
        SC  & 79.60 & 8.00 & 86.30 & 8.00 \\
        & ASC  & 79.10 & \textbf{4.35} & 85.90 & \underline{2.34} \\
        & ESC  & 79.60 & 6.06 & 86.30 & 4.69 \\
        & \modelname-SC  & \textbf{81.80} & 8.00 & \textbf{87.20} & 8.00 \\
        & \modelname-ES & \underline{80.50} &\underline{4.48} & \textbf{87.20} & \textbf{2.14} \\
        \midrule
        \multirow{5}{*}{Qwen2.5} &
        SC  & 85.00 & 8.00 & 89.50 & 8.00 \\
        & ASC  & 84.10 & \textbf{3.79} & 88.70 & \textbf{2.23} \\
        & ESC & 84.70 & 5.08 & 89.30 & 4.32 \\
        & \modelname-SC & \textbf{88.30} & 8.00 & \textbf{90.30} & 8.00 \\
        & \modelname-ES & \underline{87.70} & \underline{3.87} & \underline{90.10} & \underline{2.47} \\
        \midrule
        \bottomrule
      \end{tabular}
    }
    
    \captionof{table}{\textbf{Performance comparison for test-time scaling with sampling budget $T=8$.} 
    All the results are reported in percentage (\%).
    The best results are \textbf{bolded}. 
    The second-best is \underline{underlined}.
    }
    \label{tab:tts_8}
  \end{minipage}%
  \hfill
  \begin{minipage}[t]{0.48\textwidth}
    \centering
    \captionsetup{type=figure}
    \includegraphics[width=\linewidth]{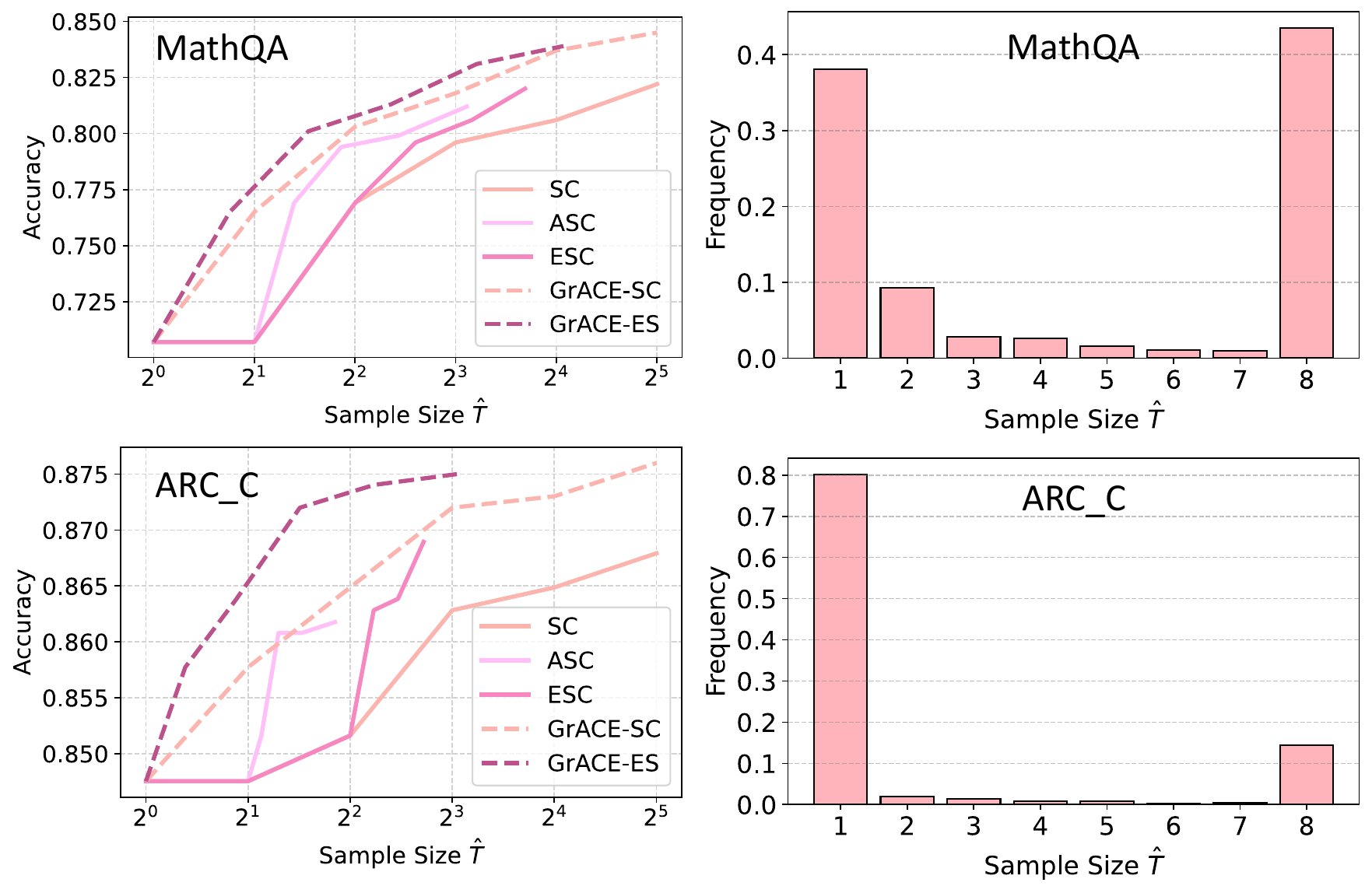}
   
    \captionof{figure}{\textbf{(Left)} Accuracy over actual sample size from $2^0$ to $2^5$. 
    \textbf{(Right)} Distribution of actual sample size $\hat{T}$ for \modelname-ES with sampling budget $T=8$, based on Llama-3.1-8B-Instruct.}
    \label{fig:tts_all}
  \end{minipage}
\vspace{-0.5cm}
\end{figure*}
\fakeparagraph{Experiment Settings.}
We use two popular question-answering datasets, TriviaQA \cite{joshi2017triviaqa} and SciQ \cite{welbl2017crowdsourcing}, for evaluating the quality of confidence produced by \modelname in the open-ended generation task.
Note that in the open-ended generation setting, the model is not provided with candidate answers, unlike in the multiple-choice setting.
Instead, we ask the model to generate answers with greedy decoding.
The correctness of model answers is determined by ROUGE score \cite{lin2004rouge} following the setting in \citet{ulmer2024calibrating}.
We select three backbone models with different architectures, scales, and training strategies:
Phi-3-3.8B \cite{abdin2024phi}, Llama-2-7B \cite{touvron2023llama}, and Llama-3.1-8B-Instruct \cite{grattafiori2024llama}, to comprehensively validate the effectiveness of our method.

We compare \modelname with  six strong confidence elicitation and calibration methods as competitors with \modelname:
\textbf{Seq. Likelihood} \cite{malininuncertainty},
\textbf{Platt Scaling} \cite{platt1999probabilistic},
\textbf{\textit{P}(True)} \cite{kadavath2022language},
\textbf{Verbal} \cite{tian2023just},
\textbf{Apricot} \cite{ulmer2024calibrating},
\textbf{ActCab} \cite{liu2024enhancing}.
We report the commonly used calibration metrics, including expected calibration error (ECE) \cite{naeini2015obtaining} and Brier score \cite{brier1950verification}, to comprehensively examine the calibration performances of different methods.
ECE measures the average discrepancy between a model’s predicted confidence and its true accuracy across discrete confidence bins.
Brier score is the mean squared difference between the predicted confidence scores and the actual correctness.
Additionally, we report AUROC to show how indicative the confidence is of the factuality of the answer.
The implementation details and detailed introduction of comparison methods are in Appendix \ref{appd:details} and Appendix \ref{appd:calibration_method}.

\fakeparagraph{\modelname produces reliable confidence in open-ended generation.}
We report our results in Table \ref{Tab: Calibration} and the corresponding reliability diagrams in Appendix \ref{appd:rd}.
\modelname outperforms one state-of-the-art method (Apricot \cite{ulmer2024calibrating}) by 3.9\%, 2.5\%, and 5.0\% in terms of ECE, Brier and AUROC.
Fine-tuning with \modelname does not affect the model utility as the accuracies on both datsets maintains at the same level with the original model.
Moreover, \modelname achieves the best AUROC among all baselines, indicating that its elicited confidence scores are highly predictive of response factuality.
On TriviaQA with Llama2-7B, ActCab yields the lowest ECE; however, as Figure \ref{fig:l-t} shows, it pushes most samples toward a confidence around 0.5, offering limited insight into answer correctness.
We find verbalized confidence calibrates the worst.
Most verbalized confidences are in the range of (0.9,1], indicating the model becomes overconfident when expressing verbalized confidence, which aligns with the conclusion of \citet{xiongcan}.
Our findings further demonstrate that LLMs learn the true distribution of prediction correctness better without any auxiliary model.
\begin{figure*}[h]
  \centering
  \vspace{-1em}
  \noindent
  \begin{minipage}[t]{0.5\textwidth}
    \centering
    \captionsetup{type=table}
    \resizebox{\linewidth}{!}{
      \begin{tabular}{c|cccc|cccc}
        \toprule \midrule
         & \multicolumn{4}{c}{\textbf{MathQA}} & \multicolumn{4}{c}{\textbf{ARC\_C}} \\
        \cmidrule(lr){2-5} \cmidrule(lr){6-9}
        \textbf{Method}& \multicolumn{1}{c}{\textit{Cali.}} & \multicolumn{1}{c}{\textit{Disc.}} & \multicolumn{2}{c}{\textit{Test-Time Scaling}} & \multicolumn{1}{c}{\textit{Cali.}} & {\textit{Disc.}} & \multicolumn{2}{c}{\textit{Test-Time Scaling}}\\
        \cmidrule(lr){2-3}
        \cmidrule(lr){4-5}
        \cmidrule(lr){6-7}
        \cmidrule(lr){8-9}
        &ECE $\downarrow$ & AUROC $\uparrow$ & Acc $\uparrow$ & $\hat{T}$ $\downarrow$ &ECE $\downarrow$& AUROC $\uparrow$ & Acc $\uparrow$ & $\hat{T}$ $\downarrow$ \\
        \midrule
        
        P(True)-SC & 33.09 & 57.14 & 79.30 & 8.00 & 42.97 & 56.19 & 86.20 & 8.00 \\
        P(True)-ES & - & - & 77.70 & 6.78 & - & - &86.00 & 6.69 \\
        Verbal-SC & \underline{22.99} & \underline{67.20} & \underline{81.10} & 8.00 &  \underline{11.20} & \underline{63.65}
        & \underline{87.00} & 8.00 \\
        Verbal-ES & - & - & 73.30 & \textbf{1.30} &  - & -& 86.40 & \textbf{1.73} \\
        \modelname-SC & \textbf{17.89} & \textbf{85.87} &  \textbf{81.80} & 8.00 & \textbf{3.12} & \textbf{87.02} & \textbf{87.20} & 8.00 \\
        \modelname-ES & - & - & 80.50 & \underline{4.48} & - & - & \textbf{87.20}  & \underline{2.14} \\

        
        \midrule
        \bottomrule
      \end{tabular}
    }
    \captionof{table}{\textbf{Performance comparison for test-time scaling with different confidence estimations for sampling budget $T=8$.} 
    All the results are reported in percentage (\%).
    The best results are \textbf{bolded}. 
    The second-best is \underline{underlined}.
    }
    \label{tab:tts_conf}
  \end{minipage}%
  \hfill
  \begin{minipage}[t]{0.48\textwidth}
    \centering
    \captionsetup{type=figure}
    \includegraphics[width=\linewidth]{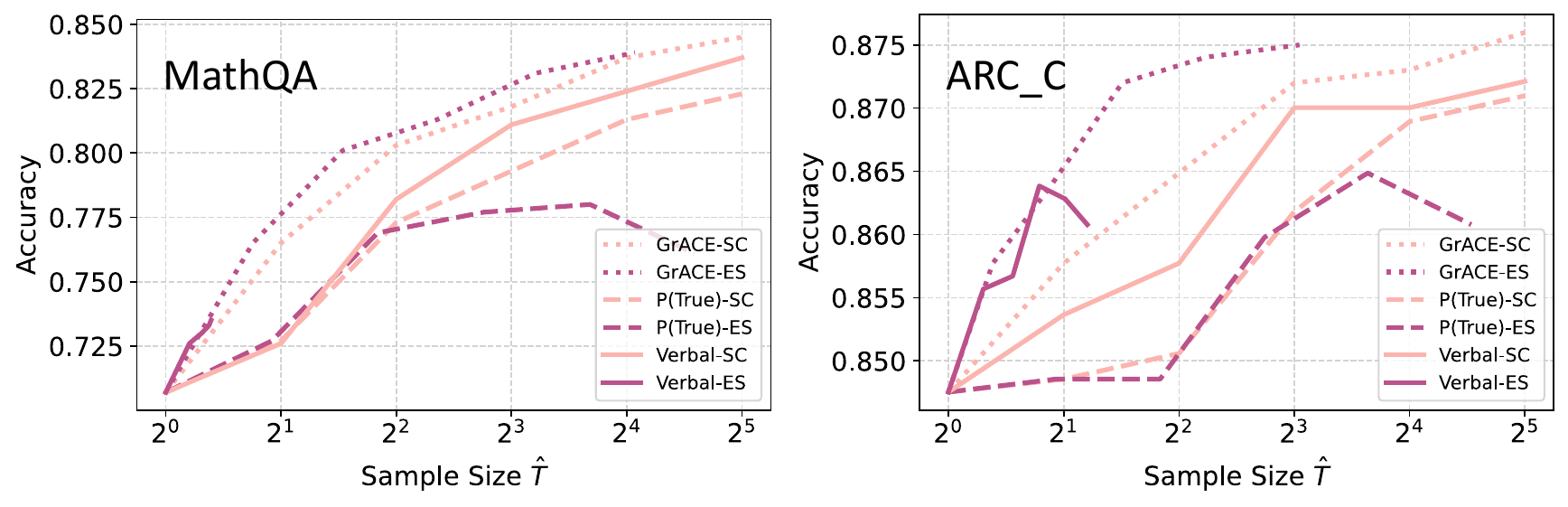}
    \captionof{figure}{\textbf{Accuracy over actual sample size $\hat{T}$ with sampling budget $T$ from $2^0$ to $2^5$.}
    The results compare adaptive test-time
    scaling strategies incorprating different confidence estimations. }
    \label{fig:tts_conf}
  \end{minipage}
\vspace{-0.5cm}
\end{figure*}

\fakeparagraph{Generalization to unseen domain.}
We study the generalization ability of confidence elicitation methods that need training (i.e., Apricot, ActCab, GrACE).
We conduct cross-validation using TriviaQA and SciQ, that is, train the model on one dataset and test on another to find out if the method performs consistently well in both in-domain (IND) and out-of-domain (OOD) setting.

The results are reported in Figure~\ref{fig:generalization}.
\modelname\ exhibits the strongest generalization, with only marginal increases in ECE (+3.3\%) and Brier score (+2.4\%), and a slight AUROC drop (-3.9\%) under domain shift.
Moreover, it achieves the highest absolute performance in the OOD setting, outperforming all baselines by a large margin.
Apricot is prone to domain shift as the AUROC decreases by 16.0\% on average, indicating that much reliance on the question semantics provides a shortcut towards calibration with limitations in generalization.
ActCab only trains a linear probe as a calibrator, which is not capable of fully capturing the correspondence between questions and correct answers.
In contrast, \modelname teaches LLM itself the true distribution of the prediction correctness, explicitly encouraging the LLM to internalize the mapping between quesion-answer pairs and the accuracy. 
This enables \modelname to generalize more effectively across domains with different semantics.
Moreover, we conduct an ablation study in Appendix \ref{appd:ablation}.

\vspace{-0.2cm}
\subsection{\modelname boosts Test-Time Scaling}
\vspace{-0.2cm}
\label{sec: tts}
\fakeparagraph{Experiment Settings.}
We evaluate how \modelname helps TTS with two widely-used reasoning LLMs: Llama-3.1-8B-Instruct \cite{grattafiori2024llama} and Qwen2.5-7B-Instruct \cite{yang2024qwen2} as they are better at instruction-following, which is essential to ensure steady performance in the test-time scaling.
The experiment is conducted on two challenging benchmark datasets: MathQA \cite{amini2019mathqa} and ARC\_Challenge (ARC\_C) \cite{clark2018think}.
We compare \modelname-SC and \modelname-ES with original Self-Consistency (\textbf{SC}) \cite{wang2022self} and two adaptive TTS methods: Adaptive Self-Consistency (\textbf{ASC}) \cite{aggarwal2023let}, Early-Stopping Self-Consistency (\textbf{ESC}) \cite{liescape}.
The details of implementation and comparison methods are in Appendix \ref{appd:details} and Appendix \ref{appd:tts_method}.

\fakeparagraph{Confidence improves consistency-based strategy.}
We show the experiment results for TTS when the sampling budget is 8 in Table \ref{tab:tts_8}.
Moreover, we evaluate the performance of all the methods with sampling budgets ranging from $2^0$ to $2^5$ and plot the results in Figure \ref{fig:tts_all}.
We observe two advantages of \modelname when applying to TTS.
\textbf{\textit{i)} \modelname improves the accuracy of vanilla self-consistency.}
As shown in Figure \ref{fig:tts_all}, \modelname-SC and \modelname-ES consistently outperform the methods based on frequency over a wide range of sample budgets.
As we examine a fixed sample budget (e.g., 8), we find that  \modelname-SC improves the accuracy over SC on MathQA and ARC\_C by 2.6\%/0.9\% with Llama3.1-8B-Instruct and 3.3\%/0.8\% with Qwen2.5-7B-Instruct, respectively.
It demonstrates that there is a gap between consistency and calibrated confidence.
We provide a case study for this phenomenon in Appendix \ref{appd:case}.
Therefore, incorporating a good confidence score as a voting weight is beneficial for making decision from samples in TTS.
\textbf{\textit{ii)} \modelname-ES improves the efficiency of TTS while preserving, even enhancing performance of the TTS.}
As shown in Figure \ref{fig:tts_all}, \modelname-ES improves the accuracy in all settings with a great reduction of sample size. 
There are 38.6\% of queries stopping sampling at the first sample whose confidence reaches the threshold for MathQA and 
80.1\% of the queries only need to sample once in the less challenging benchmark ARC\_C.
These results demonstrate that \modelname enhances both the efficacy and efficiency of TTS, helping to avoid self-consistency errors \cite{tan2025too}.

\fakeparagraph{\modelname offers the best confidence for Test-Time Scaling.}
We examine how different confidence elicitation methods affect TTS by incorprating two representative confidence elicitation methods \textit{P}(True) and Verbal into the weighted self-consistency and early-stopping strategy proposed in \seccref{subsec:tts} and test the methods with Llama-3.1-8B-Instruct.
The results are reported in Table \ref{tab:tts_conf} and Figure \ref{fig:tts_conf}.
We observe that \modelname provides the best confidence estimation for long generation, such as Chain-of-Thought, and improves self-consistency more effectively, demonstrated by the highest accuracy when applied to self-consistency while the least calibrated confidence \textit{P}(True) performs worst in TTS.
Moreover, the efficiency of early-stopping strategy is influenced by the confidence calibration.
With the same threshold, overconfident estimates (Verbal) often lead to premature termination on incorrect responses, whereas underconfident scores (\textit{P}(True)) tend to exhaust more samples and approach the sampling budget.
A well-calibrated confidence estimate serves as a reliable indicator for determining early-stopping thresholds that meet the desired accuracy.
\vspace{-0.2cm}
\section{Conclusion}
\label{sec:conclusion}
\vspace{-0.2cm}
In this paper, we propose \modelname, an approach to eliciting calibrated generative confidence from LLMs.
Unlike post-generation and verbalized confidence elicitation methods, \modelname helps the model to output calibrated high-fidelity confidence scores on-the-fly.
The LLM is trained to express its confidence by internally measuring the similarity between its last hidden state and the embedding of a special token \token.
\modelname first constructs confidence targets by connecting the uncalibrated confidence to accuracy using k-fold binning.
The model is then fine-tuned to assign a probability mass that correlates with calibration targets to the \token at the end of its response.
Extensive experiments show that \modelname outperforms six confidence elicitation methods in terms of reliability in open-ended generation, while demonstrating strong domain generalization ability.
Moreover, \modelname improves both the efficacy and efficiency of test-time scaling.

\section*{Limitations}
\label{appd:limitation}
Despite the calibration performance and extended applications, \modelname has several limitations.
First, \modelname is only able to provide a single confidence score that reflects the overall confidence for the generation.
It cannot give different confidence scores for every claim in the generation or each step in Chain-of-Thoughts, which will be addressed in our future work.
Second, \modelname focuses only on reflecting the factuality of response through confidence, while other aspects regarding model response, such as completeness, should also be considered in future work.
Finally, \modelname only assesses but does not improve the model confidence.

\section*{Ethics Statement}
Our work focus on improving the reliability of large language models.
We discourage any malicious use of our work, especially attempts to compromise LLM systems.
The artifacts and datasets in our work are all under the restriction of the license and follow the intended use.
We used GPT-5 as an AI writing assistant to refine and improve the clarity of our text.
All AI-generated suggestions were carefully reviewed and edited by the authors to ensure the integrity of the work.
The final manuscript reflects the authors’ original contributions, with AI assistance limited solely to enhancing the presentation of our findings.

\bibliography{custom}

\appendix
\clearpage
\newpage

\section{Implementation Details}
\label{appd:A}
\subsection{Prompt Template}
\label{appd:prompt}
We elaborate on the prompt templates used for open-ended generation and test-time scaling in this section.

\begin{tcolorbox}[colback=gray!5!white, colframe=gray!50!black,
                  title=Prompt Templete for Open-ended Genreation, fonttitle=\bfseries,
                  sharp corners, boxrule=0.5pt]
\texttt{Provide your best guess for the following question. Give ONLY the guess, no other words or explanation.} \\

\texttt{For example:} \\

\texttt{Question:<User's question>} \\
\texttt{Guess: <most likely guess based on the support, as short as possible; not a complete sentence, just the guess!>} \\

\texttt{Question: \{Question\}}

\end{tcolorbox}

\begin{tcolorbox}[colback=gray!5!white, colframe=gray!50!black,
                  title=Prompt Templete for Test-Time Scaling, fonttitle=\bfseries,
                  sharp corners, boxrule=0.5pt]
\texttt{For the following question, provide a step-by-step explanation of your thought process. \\
Use the format demonstrated below for your response. \\}

\texttt{```Example Format: \\
Explanation: <Your detailed explanation here, outlining how you arrived at your answer.>} \\

\texttt{Answer: <Insert your concise answer here, which should include a \{answer\_type\}> \\
Ensure that your response strictly adheres to this format. Explicitly include the words 'Explanation:', 'Answer:'.}

\end{tcolorbox}

\subsection{Experiment Settings and Hyperparameter Selection}
\label{appd:details}
In the open-ended generation calibration task, we randomly sample 2000 queries from the training data in the benchmarks for training \modelname and sample 1000 queries from the test data for evaluation.
We train \modelname with a learning rate of $1 \times 10^{-5}$ for 3 epochs, using the AdamW optimizer.
The batch size for training is 8.
We apply LoRA with rank $r=16$, scaling factor $\alpha=16$, and dropout factor of 0.05.
The LoRA module is applied to the MLP block for both backbone models.
The trainable parameters are negligible compared with the model parameters, accounting for only 0.41\%, 0.34\%, and 0.32\% of Phi3-3.8B's, Llama2-7B's, and Llama3.1-8B-Instruct's total parameters, respectively.
The threshold for ROUGE score is set to 0.3 following Ulmer et.al \cite{ulmer2024calibrating}.
The experiment is conducted on two NVIDIA GeForce RTX 3090 GPUs.

In the test-time scaling task, we randomly sample 2000 queries from the training data in the benchmarks for training \modelname and sample 1000 queries from the test data for evaluation.
To align the training process of \modelname with the implementation of TTS, we generate 16 different responses for each query in the training data with temperature $t=0.8$.
The trainable parameters account for 0.17\% and 0.13\% of Llama-3.1-8B-Instruct's and Qwen2.5-7B-Instruct's total parameters, respectively.
The model is trained with a learning rate of $lr=5\times10^{-5}$ for 3 epochs with AdamW optimizer.
The batch size is set to 4.
We apply LoRA with rank $r=32$, scaling factor $\alpha=16$, and dropout factor of 0.05.
The LoRA adapter is applied to the query and value weight of attention blocks.
The threshold for the early-stopping strategy is set to 0.8.
The experiment is conducted on two  NVIDIA A100 80GB GPUs.

We ensure the training data is equally distributed in all the confidence intervals to prevent overfitting on any specific confidence level.
This strategy helps stabilize the training process and reduce the bias for confidence elicitaion.
The hyperparameter $\gamma$ in the loss function is set to 0.1 for all the experiments.
We report the average performance over 5 random seeds.
The artifacts used in our work are all
under the restriction of the license and follow the intended use.

\subsection{Comparison Methods}

\subsubsection{Comparison Methods for Open-end Generation Calibration}
\label{appd:calibration_method}

\fakeparagraph{Sequence Likelihood} \cite{malininuncertainty} uses the length-normalized likelihood of generated sequences as predicted confidence.

\fakeparagraph{Platt Scaling}  \cite{platt1999probabilistic} fits the Seq. Likelihood to minimize the mean squared error on the validation set with scalers $a,b \in \mathbb{R}$.

\fakeparagraph{\textit{P}(True)} \cite{kadavath2022language} prompts the LLMs to decide whether or not the generation is true and uses the probability of true as the predictive confidence.

\fakeparagraph{Verbal} \cite{tian2023just} directly asks the model to assess its confidence in percentage.

\fakeparagraph{Apricot} \cite{ulmer2024calibrating} sets the accuracy of semantic clusters as the calibration target and trains an auxiliary model on the generation to predict confidence.

\fakeparagraph{ActCab} \cite{liu2024enhancing} trains a linear probe with the activation from LLMs as the calibrator.
Note that Platt Scaling, Apricot, and ActCab need to train auxiliary models while other comparison methods are training-free.

\subsubsection{Comparison Mehthods for TTS}
\label{appd:tts_method}
\fakeparagraph{SC} \cite{wang2022self} takes the majority vote from samples generated in TTS to decide the final answer.

\fakeparagraph{ASC} \cite{aggarwal2023let} calculates the cumulative frequency of each answer in the TTS process and terminates sampling when the response's relative frequency reaches the threshold.

\fakeparagraph{ESC} \cite{liescape} divides the sampling steps into windows and keep sampling until the samples within the window reach high agreement.

\subsection{Pseudocode for \modelname-ES}
\label{appd:code}
We provide the pseudocode for \modelname-ES in Algorithm \ref{code}.

\begin{algorithm}[H]
\label{code}
\caption{\modelname-ES}
\begin{algorithmic}[1]
\Require Input $\bm{x}$, Confidence Threshold $\tau$, Sample Budget $T$
\State Initialize answer list $\mathcal{A} \gets [\,]$
\State Initialize confidence list $\mathcal{C} \gets [\,]$
\For{$i = 1$ to $T$}
    \State Sample answer and confidence $\bm{y}_i, a_i, c_i \gets$ Model$(\bm{x})$
    \State Append $a_i$ to $\mathcal{A}$, $c_i$ to $\mathcal{C}$
    \If{$c_i \geq \tau$}
        \State \Return $a_i$ \Comment{Early stop}
    \EndIf
\EndFor
\Comment{Use weighted voting if threshold was never met}
\State \modelname-SC:
\State Choose $a^*_{es} = a^*_{sc}$ according to equation \ref{eq:sc}
\State \Return $a^*_{es}$
\end{algorithmic}
\label{code}
\end{algorithm}

\section{Discussions and Additional Results}

\subsection{Target Distribution}
\label{appd:target}
\begin{figure}[h]
    \centering
    \resizebox{0.5\textwidth}{!}{    \includegraphics{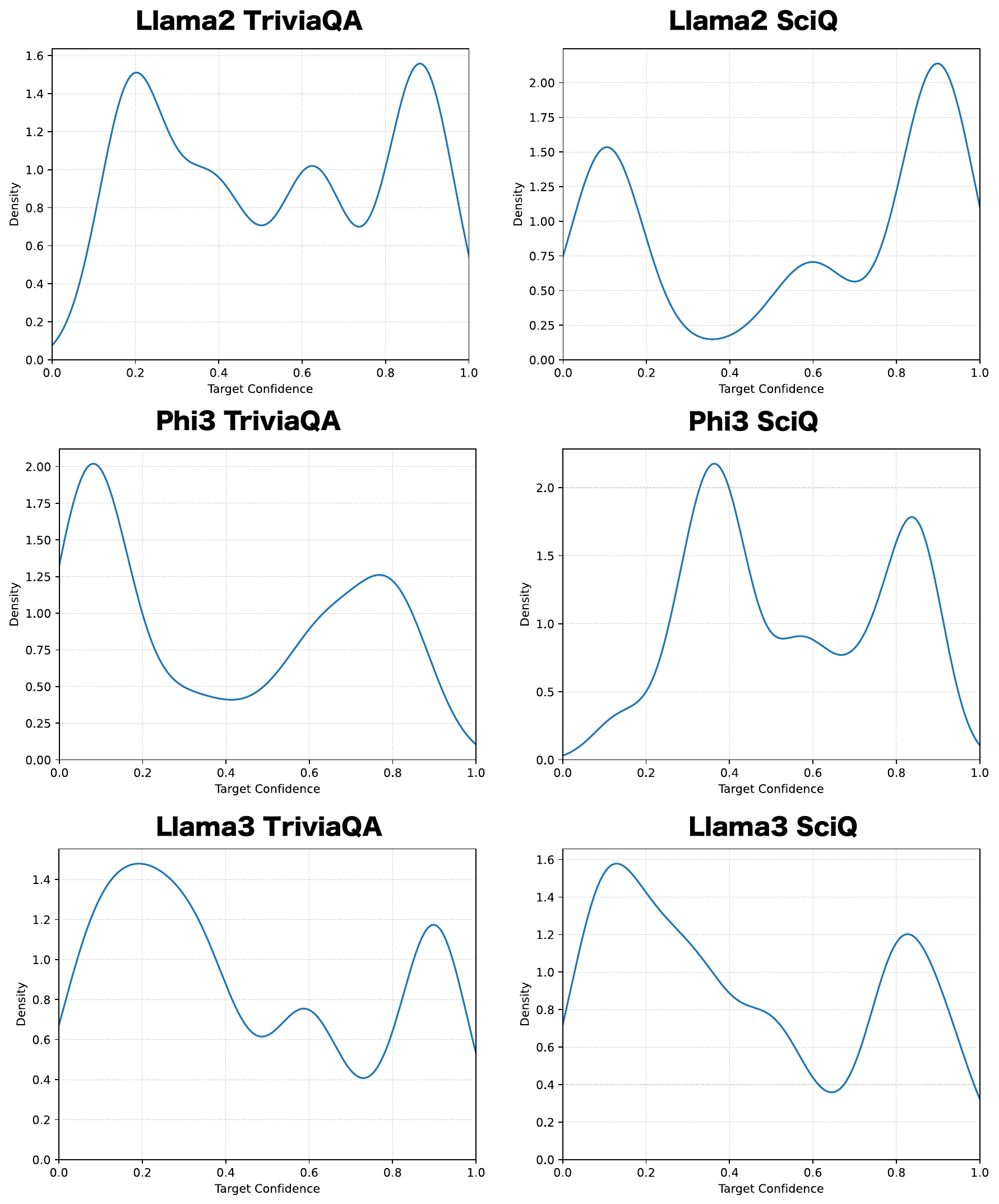} 
    }
	\caption{
    \textbf{The distribution of confidence targets.} 
	}
    
    \label{fig:target}
\end{figure}

We plot the distribution of target confidence in Figure \ref{fig:target}.
We observe that the distribution of target confidence is not concentrated around a single value. 
Instead, it exhibits a sharp distribution across the entire range of [0, 1], with relatively higher densities at both the lower and upper ends. 
Such a distribution suggests an enhanced capacity for discriminativeness.

\subsection{Relationship between the target confidence and calibration}
\label{appd:relationship}
Given the question-answer pair $(\bm{x}_i, \bm{y}_i)$, the calculation of expected calibration error (ECE) is formulated as:
\begin{equation}   \sum_{m=1}^{M}\frac{|\mathcal{B}_m|}{N}|\mathrm{Acc}(m)-\frac{1}{|\mathcal{B}_m|}\sum_{i\in \mathcal{B}_m}^{}c_i|, 
\end{equation}
where $\mathcal{B}_m$ denotes the set of indices belonging to bin $m$, $c_i$ is the confidence estimation associated with the model response $\bm{y}_i$.
Recalling that the target confidence $t_i$ in Section \seccref{subsec:cali_target} is the empirical accuracy within bin $m$, reducing the gap between the confidence $c_i$ and target confidence $t_i$ is thus equvalent to minimizing the ECE.
Therefore, optimizing loss $\mathcal{L}_C$ in Eq. \ref{eq:loss} directly contributes to improving model calibration.

\subsection{Reliability Diagrams}
\label{appd:rd}
\begin{figure*}[!h]
\centering

\includegraphics[width=\linewidth]{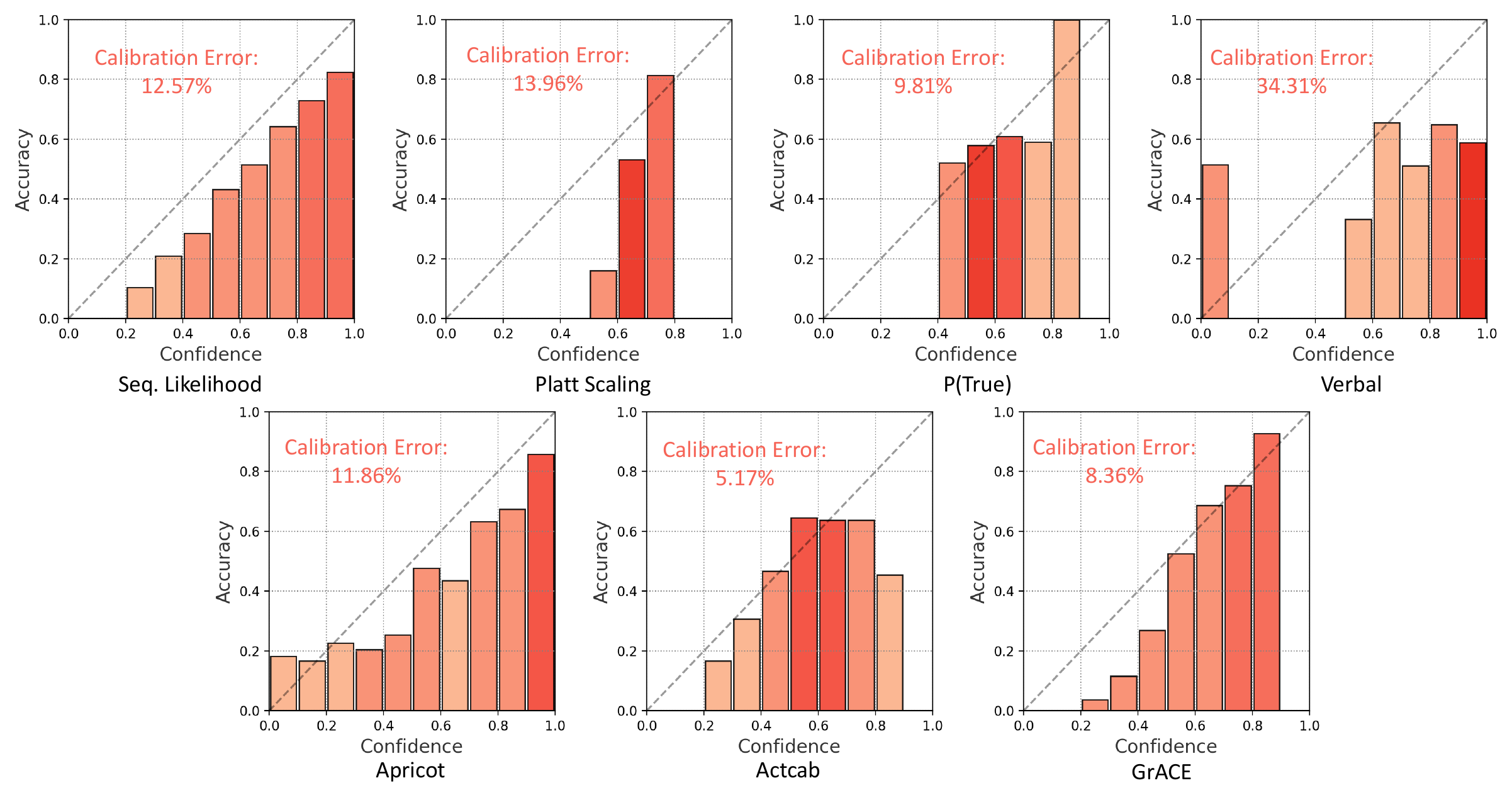} 

\caption{\textbf{The reliability diagram of various confidence elicitation methods applied to Llama2-7B on the TriviaQA test dataset.}
The color indicates the proportion of total responses
contained in each bin.
The color close to red suggests a larger value.}

\label{fig:l-t}
\end{figure*}

\begin{figure*}[!h]
\centering
\includegraphics[width=\linewidth]{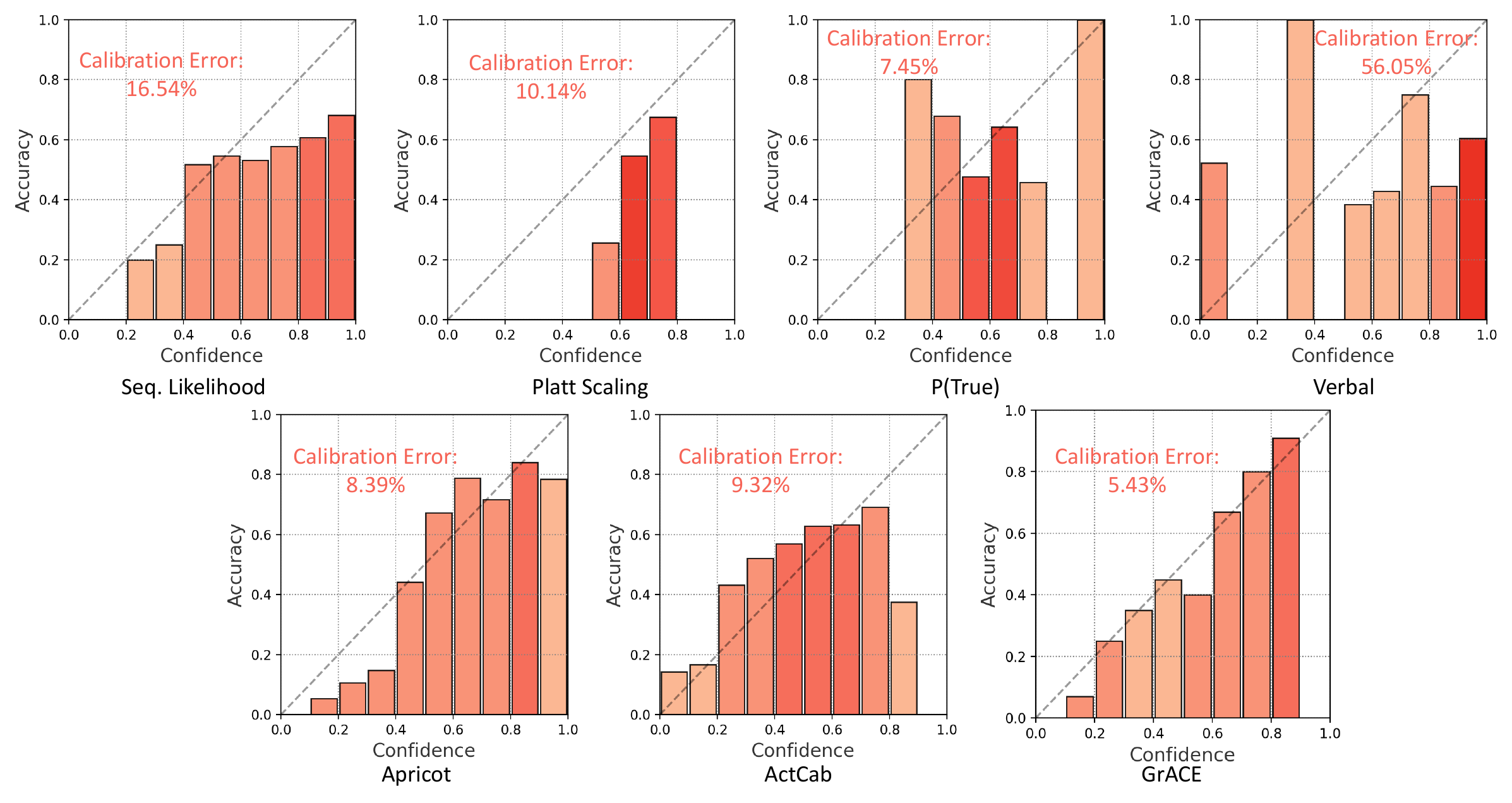} 
\caption{\textbf{The reliability diagram of various confidence elicitation methods applied to Llama2-7B on the SciQ test dataset.}
The color indicates the proportion of total responses
contained in each bin.
The color close to red suggests larger value.}
\label{fig:l-s}
\end{figure*}

\begin{figure*}[!h]
\centering

\includegraphics[width=\linewidth]{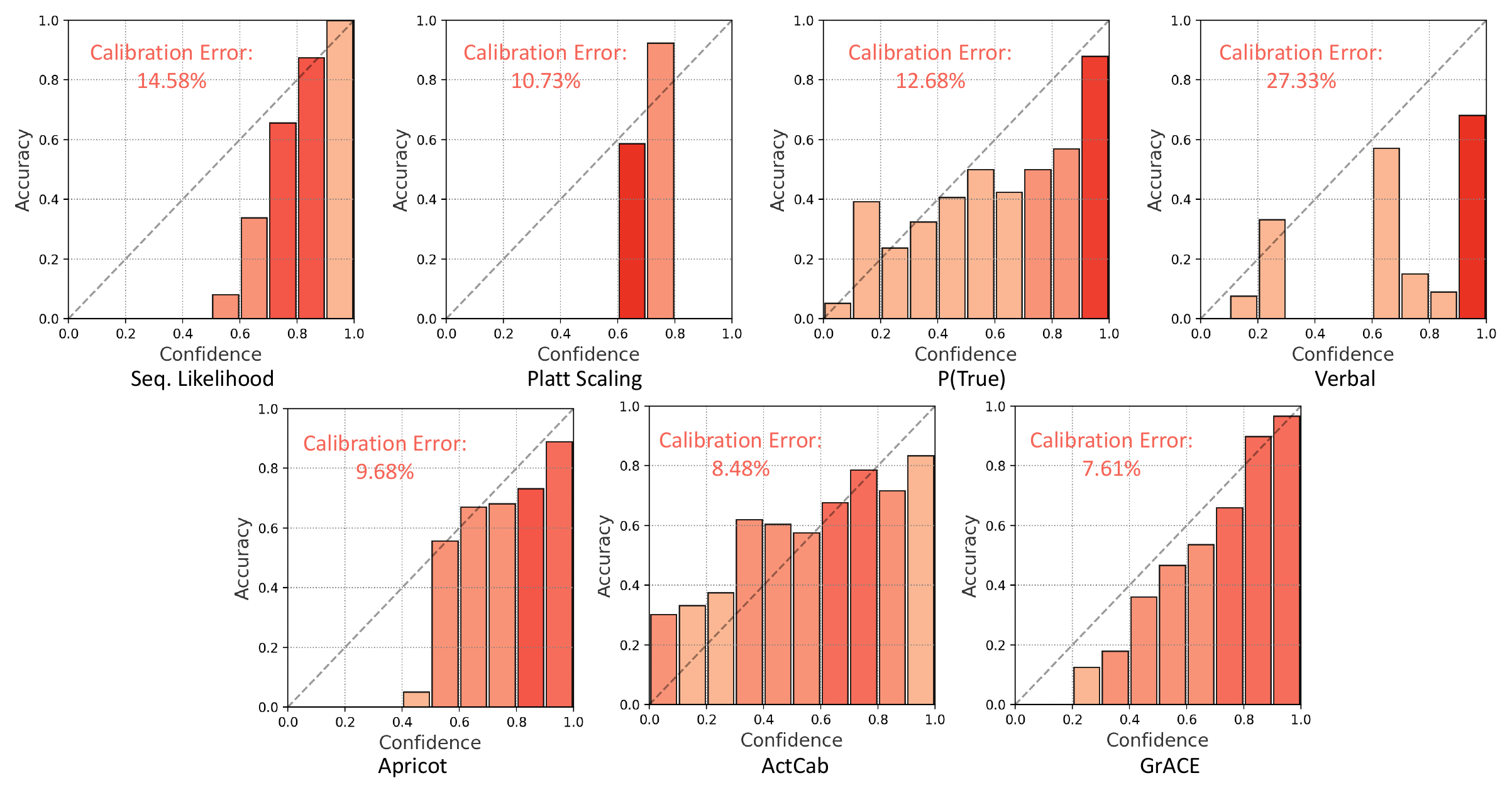} 

\caption{\textbf{The reliability diagram of various confidence elicitation methods applied to Phi3-3.8B on the TriviaQA test dataset.}
The color indicates the proportion of total responses
contained in each bin.
The color close to red suggests larger value.}

\label{fig:p-t}
\end{figure*}

\begin{figure*}[!h]
\centering

\includegraphics[width=\linewidth]{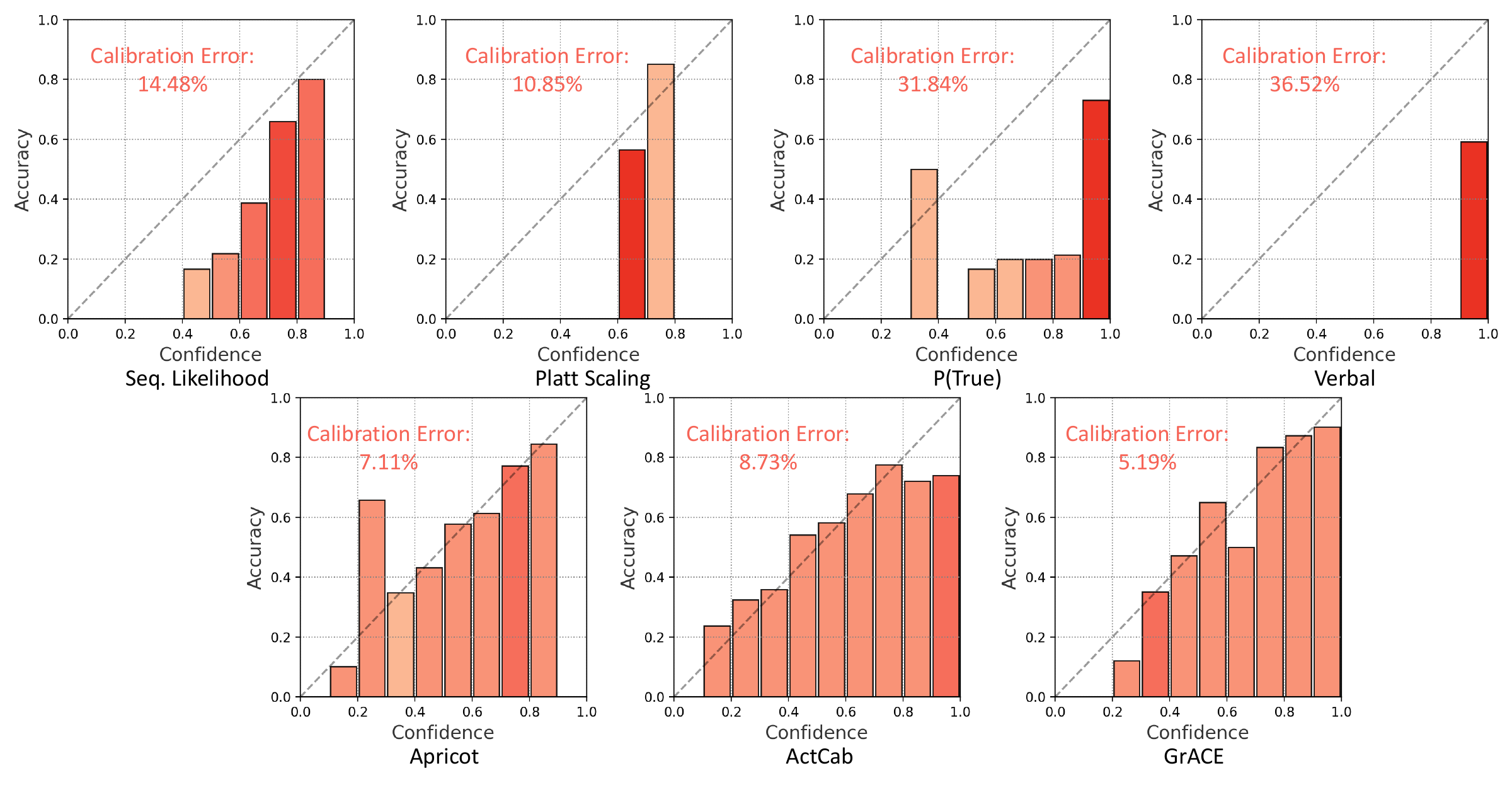} 

\caption{\textbf{The reliability diagram of various confidence elicitation methods applied to Phi3-3.8B on the SciQ test dataset.}
The color indicates the proportion of total responses
contained in each bin.
The color close to red suggests larger value.}

\label{fig:p-s}
\end{figure*}

\begin{figure*}[!h]
\centering
\includegraphics[width=\linewidth]{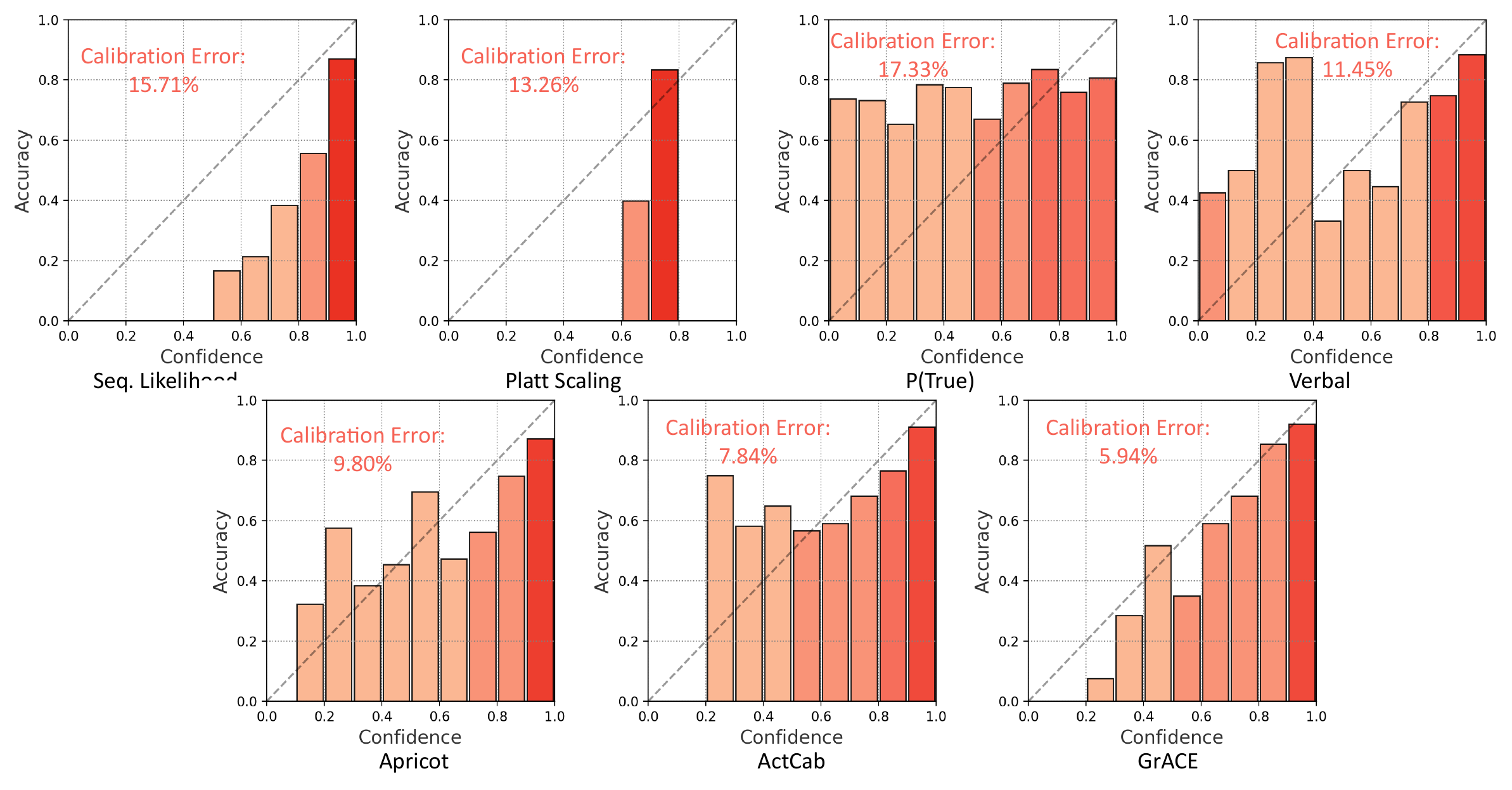} 
\caption{\textbf{The reliability diagram of various confidence elicitation methods applied to Llama3.1-8B-Instruct on the TriviaQA test dataset.}
The color indicates the proportion of total responses
contained in each bin.
The color close to red suggests larger value.}
\label{fig:l3-t}
\end{figure*}

\begin{figure*}[!h]
\centering
\includegraphics[width=\linewidth]{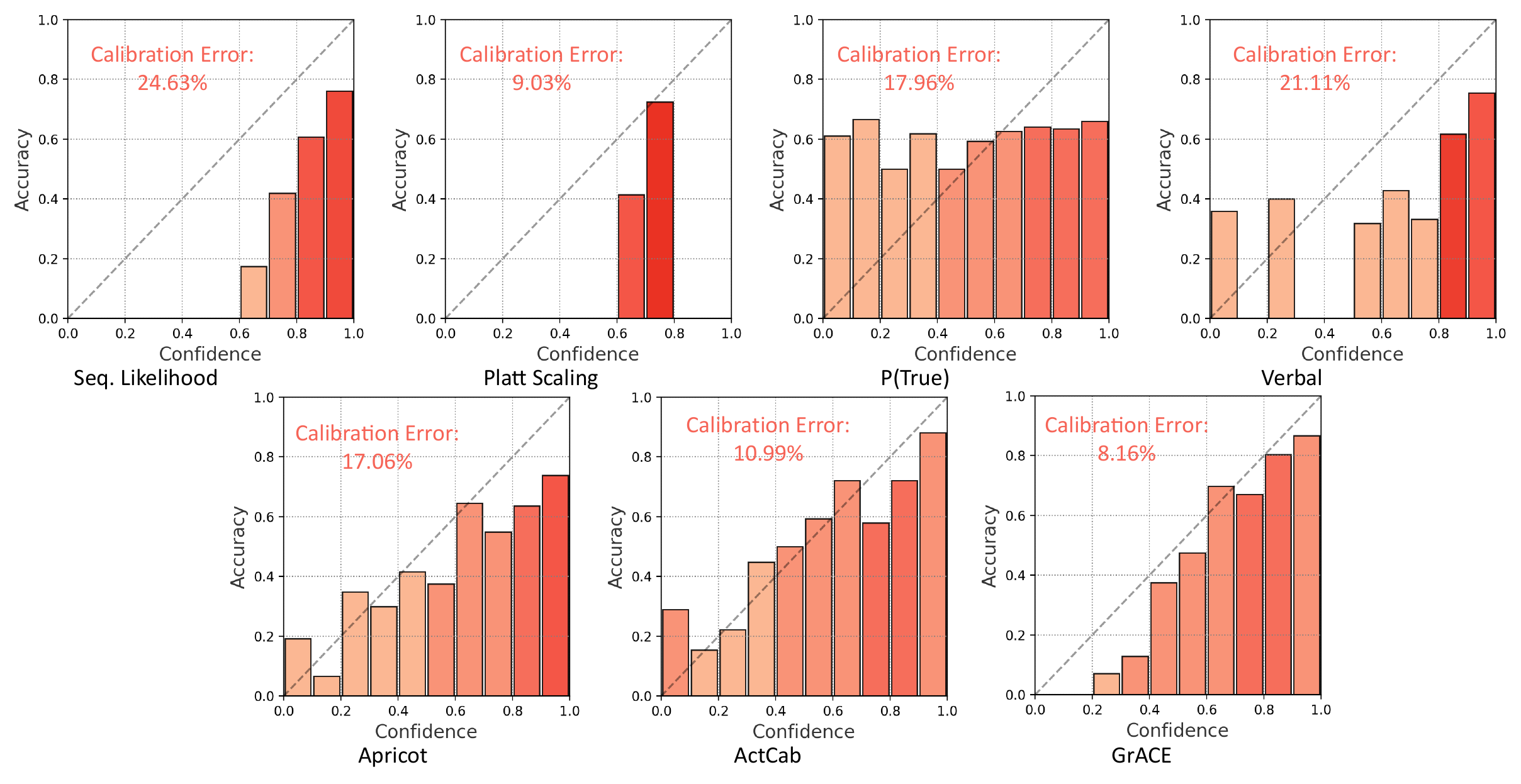} 
\caption{\textbf{The reliability diagram of various confidence elicitation methods applied to Llama3.1-8B-Instruct on the SciQ test dataset.}
The color indicates the proportion of total responses
contained in each bin.
The color close to red suggests larger value.}
\label{fig:l3-s}
\end{figure*}

The reliability diagrams for the calibration of different confidence elicitation methods are shown in Figure \ref{fig:l-t} (Llama2-7B on TriviaQA), Figure \ref{fig:l-s} (Llama2-7B on SciQ), Figure \ref{fig:p-t} (Phi3-3.8B on TriviaQA), Figure \ref{fig:p-s} (Phi3-3.8B on SciQ), Figure \ref{fig:l3-t} (Llama3.1-8B-Instruct on TriviaQA), and Figure \ref{fig:l3-s} (Llama3.1-8B-Instruct on SciQ).

\subsection{Ablation Study}
\label{appd:ablation}
We conduct an ablation study using three variants of \modelname to investigate the contributions of different components, evaluated on the TriviaQA dataset with Llama2-7B.
\textbf{\modelname-Disc} expresses confidence with discrete tokens. 
In the training stage, each data point is assigned to one of $N=10$ bins as described in \seccref{subsec:cali_target}.
The bins are represented by a set of special tokens \{\texttt{<B1>}, \ldots, \texttt{<BN>}\}, which are appended to the end of the response. 
The model is trained to predict the correct token using the $\mathcal{L}_{\mathrm{SFT}}$.
During evaluation, each bin corresponds to a representative confidence score from the set $\{0.05, 0.15, \ldots, 0.95\}$ , which approximates the midpoint of each confidence interval. 
\textbf{\modelname-Pre} places the \token immediately after the input prompt rather than at the end of the response, thereby eliciting confidence based solely on the question context.
\textbf{\modelname-SFT} includes the \token in the calculation of $\mathcal{L}_{\mathrm{SFT}}$ when calculating Eq. \ref{eq:loss}, explicitly training the model to allocate probability mass to it.

The results are reported in Table \ref{tab:ablation}.
\modelname-Disc leads to a 1.93\% increase in ECE and a 7.01\% drop in AUROC, indicating that a discretized representation of confidence is insufficiently precise for reliably reflecting response quality.
Although \modelname-Disc shares similarity with verbalized confidence methods which give a discrete expression of confidence, it performs better in terms of calibration because the new tokens are specifically trained to express confidence associated with factuality, while verbalized confidence methods are interfered with natural language generation.
GrACE-Pre drops by 28.55\% in AUROC, suggesting that the confidence scores derived solely from question information are not reliable indicators of the correctness of the model response.
GrACE-SFT suffers a great increase in ECE (+4.39\%) but a rather trivial decrease in AUROC (-2.98\%), because optimizing \token with $\mathcal{L}_\mathrm{SFT}$ puts extra probability mass on \token, leading to overconfidence.
However, the relatively small drop in AUROC indicates that, despite the increased miscalibration, the elicited confidence scores still retain a strong correlation with response actuality.
\begin{table}[ht]
\centering
\resizebox{0.45\textwidth}{!}{
    \begin{tabular}{c c c c}
\toprule
\midrule
\textbf{Method} & \textbf{ECE}$\downarrow$ & \textbf{Brier}$\downarrow$ & \textbf{AUROC}$\uparrow$ \\
\midrule
\textbf{\modelname-Disc} & 10.29& 20.23 & 77.21 \\
\textbf{\modelname-Pre} & 10.45& 26.32 & 55.67\\
\textbf{\modelname-SFT} & 12.75& 19.28 & 81.24 \\
\textbf{\modelname} & \textbf{8.36}& \textbf{16.78} & \textbf{84.22}  \\
\midrule
\bottomrule
\end{tabular}
}
\caption{\textbf{Ablation study results on Triviaqa using Llama2-7B.}
All results are reported in percentage (\%).
The best results are \textbf{bolded}.}
\label{tab:ablation}
\end{table}

\subsection{Case Study}
\label{appd:case}
We provide a case in which the query is from MathQA and the samples are generated by Llama-3.1-8B-Instruct in Table \ref{tab:case_study}.
In this case, the four samples give the consistent answer (D) to the query, which is also the correct answer.
However, the model generates the fourth sample with confidence 0.1 while the confidences are higher than 0.9 for other samples.
Taking a closer look at the samples, we find that the confident samples explicitly express the correct answer (8 days) in the thinking process.
The sample with confidence 0.1 skips the step where the specific answer is derived, leading to low confidence to the final answer.
This case demonstrates that within the samples that are consistent in final answer, there are also samples with low confidence.
\modelname does not capture the superfacial self-consistancy information but the internal confidence in the model.

\begin{table*}[!h]
\centering
\renewcommand\arraystretch{0.8}
\begin{tabular}{|m{0.22\textwidth}|m{0.65\textwidth}|m{0.12\textwidth}|}
\hline
\centering\textbf{Query} & \centering\textbf{Sample} & \textbf{Confidence} \\
\hline

\multirow{4}{=}{
\newline
\newline
\newline
\newline
\newline
\newline
\newline
\newline
\newline
\newline
\newline
\newline
\newline
\newline
A contractor estimates 
that 10 people can paint Mr Khan's house in 4 days. If he uses 5 people instead of 10, how long will it take to complete the job?

Options:

A. 10

B. 4

C. 5

D. 8

E. 6
}

& Explanation: To solve this problem, we first need to understand the relationship between the number of people and the time taken to complete the job. The original information given is that 10 people can paint the house in 4 days. This means that the total amount of work done is constant.

...

New time = Total work / New number of people

New time = 40 man-days / 5 people

New time = 8 days
\newline
\newline
Therefore, if the constructor uses 5 people instead of 10, it will take them \textcolor{red}{8 days} to complete the job.
\newline
\newline
\textbf{Answer: D <CNF>}
& 0.992 \\

\cline{2-3}

& Explanation: To solve this problem, we can use the concept of work rate. The initial situation is that 10 people can paint the house in 4 days. 

...

To find out how long it will take for 5 people to complete the job, we divide the total work by the number of people: 40 person-days / 5 people = \textcolor{red}{8 days}.
\newline
\newline
\textbf{Answer: D <CNF>}
& 0.993 \\

\cline{2-3}

& Explanation: To solve this problem, we need to understand the concept of work and time. When the number of workers decreases, the time taken to complete the job increases proportionally. We can use the concept of'man-days' to solve this problem.

...

Time = Total man-days / Number of people

Time = 40 / 5

Time = 8 days
\newline
\newline
Therefore, if the constructor uses 5 people instead of 10, they will take \textcolor{red}{8 days} to complete the job.
\newline
\newline
\textbf{Answer: D <CNF>}
& 0.988 \\

\cline{2-3}

& Explanation: To solve this problem, we first need to understand the relationship between the number of people and the time it takes to complete the job. Since the number of people is being reduced from 10 to 5, it implies that the work will take longer to complete because there are fewer workers. 
The initial ratio is 10 people in 4 days. If we reduce the number of people by half (10/2 = 5), the time taken to complete the job will increase by a factor of 2, since there are fewer workers to divide the work among.
\newline
\newline
\textbf{Answer: D <CNF>}
& 0.100 \\

\hline
\end{tabular}
\caption{Examples of consistent sample with different confidences.}
\label{tab:case_study}
\end{table*}

\end{document}